\pdfoutput=1

\documentclass[11pt]{article}

\usepackage{acl}
\usepackage[affil-it]{authblk}

\usepackage{times}
\usepackage{latexsym}

\usepackage[T1]{fontenc}

\usepackage[utf8]{inputenc}

\usepackage{microtype}

\usepackage{inconsolata}

\usepackage{graphicx}

\usepackage[ruled,vlined]{algorithm2e}
\usepackage{multirow}
\usepackage{booktabs,caption}
\usepackage[flushleft]{threeparttable}

\usepackage{latexsym}

\usepackage{microtype}
\usepackage{amsmath, xparse}
\usepackage{makecell}
\usepackage{ragged2e}
\usepackage{xcolor}
\usepackage{graphicx}
\usepackage{subfigure}
\usepackage{amssymb}
\usepackage{pifont}
\usepackage{array}
\usepackage{appendix}
\newcommand{\xmark}{\ding{55}}%
\newcommand{\cmark}{\ding{51}}%

\title{VTechAGP: An Academic-to-General-Audience Text Paraphrase Dataset and Benchmark Models}

\author{\bf Ming Cheng$^{1*}$, Jiaying Gong$^{1}$\thanks{Ming cheng and Jiaying Gong contributed equally.}, Chenhan Yuan$^{2}$, \\ \bf William A. Ingram$^{1}$,  Edward Fox$^{1}$, Hoda Eldardiry$^{1}$ \\
$^{1}$Virginia Tech, $^{2}$University of Manchester \\
\texttt{\{ming98,gjiaying,waingram,fox,hdardiry\}@vt.edu}, chenhan.yuan@manchester.ac.uk
}

\begin{document}
\maketitle
\begin{abstract}
Existing text simplification or paraphrase datasets mainly focus on sentence-level text generation in a general domain. These datasets are typically developed without using domain knowledge.
In this paper, we release a novel dataset, \textbf{VTechAGP}, which is the first \textit{academic-to-general-audience} text paraphrase dataset consisting of document-level thesis and dissertation academic and general-audience abstract pairs from 8 colleges authored over 25 years. We also propose a novel \textit{dynamic soft prompt} generative language model, \textbf{DSPT5}, for the academic-to-general-audience text paraphrasing task. For training, we leverage a contrastive-generative loss function to learn the keyword vectors in the dynamic prompt. For inference, we adopt a crowd-sampling decoding strategy at both semantic and structural levels to further select the best output candidate. 
We evaluate \textbf{DSPT5} and various state-of-the-art large language models (LLMs) from multiple perspectives.
Results demonstrate that the SOTA LLMs do not provide satisfactory outcomes, while the lightweight \textbf{DSPT5} can achieve competitive results.
To the best of our knowledge, we are the first to build a benchmark dataset and solutions for academic-to-general-audience text paraphrase dataset~\footnote{\href{https://github.com/waingram/VTechAGP-Dataset}{https://github.com/waingram/VTechAGP-Dataset}, Dataset: \href{ https://doi.org/10.5281/zenodo.14833932}{ https://doi.org/10.5281/zenodo.14833932}}. 
\end{abstract}

\section{Introduction}
Text generation aims to produce understandable text in human language from various sources of input data.
Among them, text-to-text generation remains an important and challenging task with extensive applications such as language translation~\cite{ranathunga2023neural, dabre2020survey}, paraphrase generation~\cite{singh2022paraphrase}, text simplification~\cite{martin2023review}, etc. 

Existing text simplification and paraphrase generation datasets (shown in Table~\ref{tab:example} in the Appendix) mainly focus on sentence-level translation. The recent paragraph-level dataset WikiAuto~\cite{acl/JiangMLZX20}, derived from WikiLarge~\cite{zhang-lapata-2017-sentence} and Newsela~\cite{xu-etal-2015-problems}, still suffers from a lack of domain diversity and specification, which are sourced from Wikipedia and news articles.
Besides, the objective of these studies on the aforementioned datasets is to simplify the text for children at lower grade levels.
Furthermore, existing works on lay summarization involve brief summarization limited to medicine and biological domains~\cite{devaraj-etal-2021-paragraph, flores-etal-2023-medical, jiang2024llm, zaman7sats}.
However, our proposed VTechAGP involves translation from technical language to general-audience language across a broader set of multiple domains.
We aim to generate text from the domain-specific to a general level of understanding while keeping it scientifically accurate and easy to understand, in order to encourage and facilitate interdisciplinary collaboration across different research fields.

In this paper, we pioneer the research of academic-to-general-audience text generation by introducing a new benchmark dataset VTechAGP, which is derived from electronic theses and dissertations (ETDs) at Virginia Tech over twenty-five years.
VTechAGP consists of document-level abstract pairs (academic abstract and general-audience abstract). 
VTechAGP also provides other information such as title, discipline, degree level, etc.
This auxiliary information shows the potential of VTechAGP for other tasks such as topic generation, etc.
In addition, the abstracts in VTechAGP come from multiple domains (colleges) and VTechAGP is labeled with each specific domain and provides the domain knowledge keywords. 
More details about VTechAGP are presented in Sec.~\ref{sec:data_source} and Table~\ref{tab:dataset_columns}.

Based on VTechAGP, we evaluate several SOTA pre-trained large language models (LLMs), such as LLaMA2~\cite{touvron2023llama}, Claude2~\cite{claude}, ChatGPT~\cite{brown2020language}, etc., to establish the baseline performance.
However, these SOTA pre-trained LLMs have demonstrated the following limitations: 
(1) Some LLMs do not provide public APIs, or the APIs are not free. Also, some LLMs (e.g., Claude2) do not provide fine-tuning, making them less adaptable to specific tasks.
(2) The model size of LLMs is very large. For example, LLaMA2 has about 65 billion parameters.
Fine-tuning these LLMs is resource-intensive in terms of memory and computation time. 
Even the inference implementation requires more memory and time.
(3) The pre-trained LLMs do not show competitive performance for the academic-to-general-audience text paraphrasing task on VTechAGP in Sec.~\ref{sec:experiment}.

To address the above challenges, we propose DSPT5, a dynamic soft prompt-based generative model with the crowd sampling decoding strategy during the inference stage. 
DSPT5 is built based on the pre-trained T5~\cite{raffel2020exploring}, which has only about 220 million parameters. 
In particular, the dynamic soft prompt template in DSPT5 can automatically adapt to different academic domains by changing the keywords extracted from the academic abstract.
The prompt encoder in DSPT5 is trained to generate and fine-tune keyword vectors combined with the dynamic prompt template.
To this end, we design a hybrid loss function with generative language model loss and contrastive loss to jointly learn the generated text representations as well as the ability to distinguish technical keywords from non-technical keywords.
During inference, DSPT5 employs two alignment functions at both the semantic and structural levels to select the best candidate for the final generated output.

The contributions can be summarized as follows: 
(1) \textbf{Dataset:} We construct VTechAGP, the first academic-to-general-audience text paraphrase dataset. VTechAGP is a document-level text generation dataset with multiple technical domains.
(2) \textbf{Baselines:} We implement several SOTA LLMs as benchmarks to compare the performance with our proposed model DSPT5. Experimental results show that there is still a huge room for further improvement of the existing LLMs.
(3) \textbf{Approach:} We propose a lightweight model, DSPT5, which utilizes dynamic soft prompts with a hybrid loss function and a new crowd decoding strategy. Experimental results show that DSPT5 can achieve competitive results with SOTA LLMs.
(4) \textbf{Evaluation:} We explore various evaluation metrics for the academic-to-general-audience text paraphrasing task on VTechAGP from different perspectives, including document-level embedding-based, word-based, and end-to-end metrics. In addition, simplicity, diversity, readability, and toxicity are also considered for the performance evaluation.

\section{Related Work} 
\subsection{Text Generation Datasets}
The most commonly used datasets for text generation focus on cross-lingual machine translation, such as WMT datasets~\cite{farhad2021findings, bojar2016findings} for news translation, IWSLT datasets~\cite{scarton-etal-2019-estimating, lee-etal-2022-docmt5} for document translation of TED talks, parallel corpus datasets (Europarl~\cite{koehn2003statistical}, UN Parallel Corpus~\cite{ziemski2016united}, OPUS~\cite{zhang-etal-2020-improving}, and Tatoeba~\cite{tiedemann2020tatoeba}) for sentence-level translation.
Additional text generation datasets focus on text/sentence simplification, such as datasets from news articles~\cite{xu-etal-2015-problems, stodden-etal-2023-deplain}, clinical reports~\cite{luo-etal-2022-benchmarking}, Wikipedia~\cite{zhu-etal-2010-monolingual, xu-etal-2016-optimizing, zhang-lapata-2017-sentence, alva-manchego-etal-2020-asset, naderi2019subjective, aumiller-gertz-2022-klexikon}, and other language learning resources~\cite{vajjala-lucic-2018-onestopenglish}.
Existing paraphrase datasets include phrasal and lexical paraphrases~\cite{ganitkevitch-etal-2013-ppdb}, question pairs from Wikidata~\cite{fader2013paraphrase}, and Quora~\cite{ijcai2017p579}, the same posts shared by Twitter~\cite{lan-etal-2017-continuously}, different image captions~\cite{lin2014microsoft}, and a machine translation dataset involving back translation~\cite{huang-etal-2023-paraamr}.
However, these datasets are limited to sentence-level translation and non-academic data with a lack of domain expert knowledge.

\subsection{Text Generation Methods}
\begin{table*}[htp]
\caption{Dataset statistics of VTechAGP over eight colleges. Statistics are reported in the format of academic (source) / general audience (target) documents.}
\label{tab:dataset}
\centering
\small
\tabcolsep=0.12cm
\begin{tabular}{lcccc}
\hline
College                                 & Avg. \# Sentence & Avg. Sentence Len. & MTLD          & Readability Consensus \\ \hline
Agriculture and Life Sciences (ALS)     & 15.06 / 13.80    & 27.42 / 26.37      & 73.59 / 68.58 & 15th-16th / 14th-15th \\
Architecture, Arts, and Design (AAD)    & 10.83 / 8.99     & 26.67 / 26.07      & 68.24 / 68.76 & 13th-14th / 10th-11th \\
Engineering (ENG.)                      & 14.03 / 11.99    & 26.87 / 25.66      & 69.93 / 66.47 & 13th-14th / 14th-15th \\
Liberal Arts and Human Sciences (LAHS)  & 10.12 / 8.60     & 30.24 / 28.46      & 68.02 / 67.06 & 16th-17th / 15th-16th \\
Natural Resources and Environment (NRE) & 14.48 / 12.66    & 28.87 / 27.33      & 76.69 / 71.99 & 15th-16th / 10th-11th \\
Science (SCI.)                          & 14.19 / 11.75    & 27.35 / 25.73      & 74.90 / 68.48 & 16th-17th / 15th-16th \\
Business (BUS.)                         & 11.76 / 10.24    & 28.55 / 27.63      & 68.46 / 70.35 & 18th-19th / 17th-18th \\
Veterinary Medicine (VM)                & 15.61 / 13.16    & 26.83 / 25.92      & 77.77 / 70.15 & 16th-17th / 14th-15th \\ \hline
All                                     & 13.68 / 11.66    & 27.68 / 26.38      & 71.36 / 67.91 & 15th-16th / 13th-14th \\ \hline
\end{tabular}
\end{table*}

With the emergence of Transformers~\cite{vaswani2017attention}, pre-trained language models (PLMs) -- such as BERT~\cite{devlin-etal-2019-bert}, BART~\cite{lewis-etal-2020-bart}, ChatGPT~\cite{brown2020language}, T5~\cite{raffel2020exploring}, Claude2~\cite{claude}, and LLaMA2~\cite{touvron2023llama} -- have been developed and fine-tuned for various generation tasks.
For example, fine-tuning LLMs on a customized dataset~\cite{sun-etal-2023-teaching, yermakov-etal-2021-biomedical, Dathathri2020Plug} with prompts~\cite{luo-etal-2023-vector, wan-etal-2023-pip, yang-etal-2023-tailor, kew-etal-2023-bless, chen2023mixture} is a popular method for text generation.
Several decoding strategies (i.e, minimum Bayes decoding~\cite{suzgun-etal-2023-follow}, beam search~\cite{yoon-bak-2023-diversity, zhang-etal-2020-pointer}, probability-based sampling~\cite{li2022diffusion, guo2018long, xu2022best}, nucleus sampling~\cite{Holtzman2020The}, prefix-adaptive decoding~\cite{pei-etal-2023-preadd}, and contrastive loss~\cite{an2022cont}) are applied to select the best generated candidates.
However, these studies only focus on paraphrase generation or text simplification in a general domain.
Our focus is not on shortening the text, but rather on how these PLMs can paraphrase academic language into non-academic language while maintaining accuracy and simplicity.

\section{Dataset Construction}~\label{sec:data_source} 
\vspace{-6mm}
\subsection{Data Source and Dataset Collection}
Virginia Tech has been a leader in ETDs for more than twenty-five years. 
It was the first university to require electronic submission of ETDs, beginning in 1997. 
Virginia Tech's ETDs are accessible through VTechWorks,\footnote{Virginia Tech Electronic Theses and Dissertations: \url{https://hdl.handle.net/10919/5534}} a digital repository created through a collaboration between the Graduate School and the University Libraries. 
In the fall of 2016, the Graduate School added a new requirement for ETDs: the inclusion of a \textit{general audience abstract} in addition to the traditional academic abstract. 
The ETD submission system was updated in 2019 to include a separate field for the general audience abstract.
Since then, most ETDs have included both an academic abstract and a general audience abstract, which are captured as distinct metadata fields in VTechWorks. 
Like many institutional repositories, VTechWorks supports the Open Archives Initiative Protocol for Metadata Harvesting (OAI-PMH)~\cite{oai_pmh_2015}.
OAI-PMH is a standard protocol for retrieving metadata records from digital repositories. 
It provides a framework that enables data exchange between various systems, supporting consistent metadata formats.
To create the VTechAGP dataset, we used OAI-PMH to harvest metadata from VTechWorks. 
We specifically queried the VTechWorks OAI-PMH endpoint to retrieve metadata records for ETDs containing both academic and general audience abstracts. 
The OAI-PMH endpoint provided us with an XML record for each ETD. 
This data encompassed a range of metadata elements defined in a qualified Dublin Core schema. 
We identified specific fields necessary for our dataset and extracted the text content. 
This step involved parsing the XML structure, locating the relevant elements, and retrieving the textual data they contained.
We mapped the extracted metadata to specific columns in a CSV file. 
Each piece of extracted metadata was mapped to a corresponding column header in the CSV file: `identifier\_uri', `title', `abstract', `abstract\_general', `subject\_terms', `discipline', `department', `degree', `degree\_level', and `type'.
A description is given in Table~\ref{tab:dataset_columns} in the Appendix.
We provide the data analysis in Sec.~\ref{sec:appendix_analysis}.

\subsection{Task Definition}
\begin{figure*}[htp] 
 \center{\includegraphics[height=6.5cm,width=\textwidth]{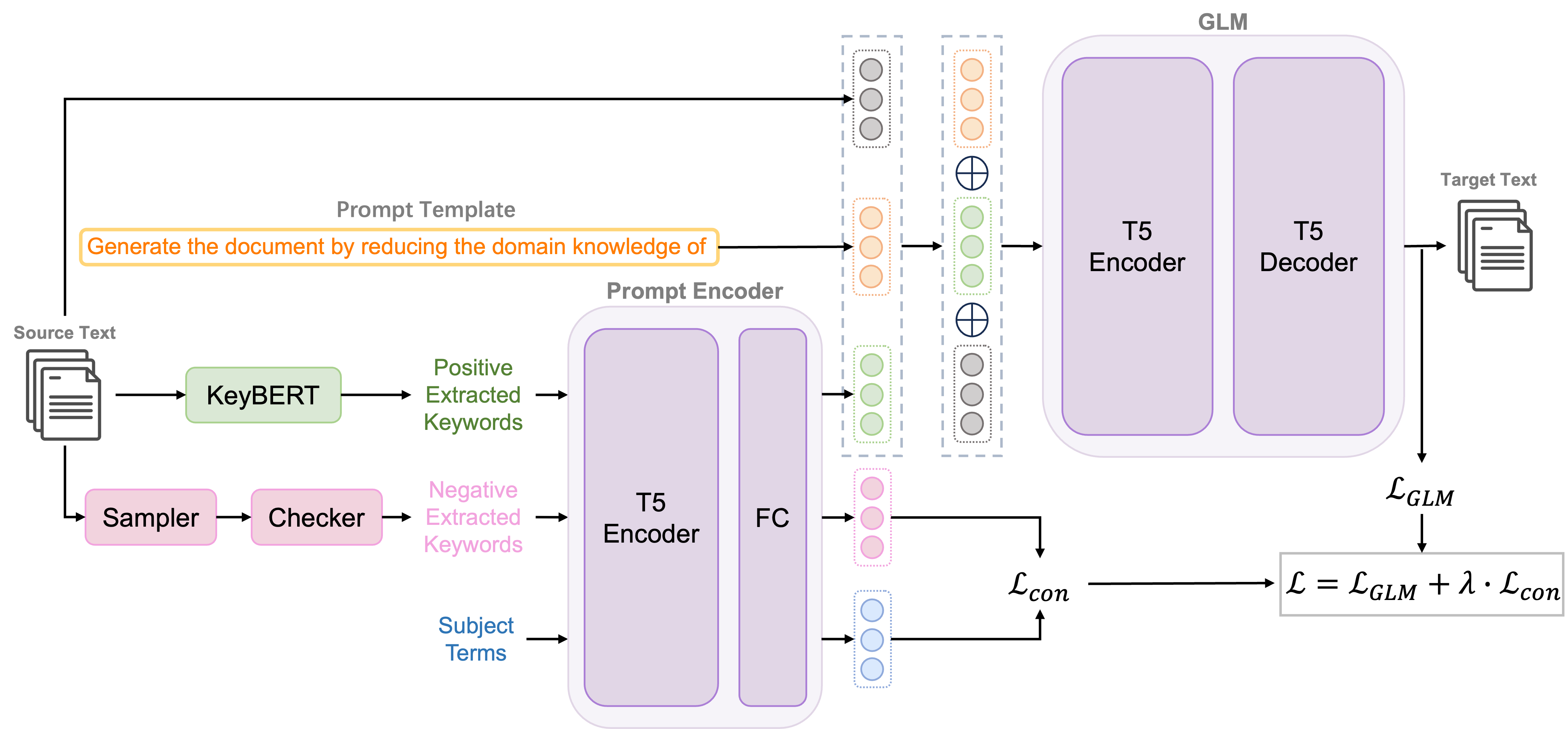}}
 \caption{\label{fig:framework} Dynamic soft prompt generation of our proposed model DSPT5. The framework includes a prompt encoder and a generative language model fine-tuning together with a hybrid loss. FC is a fully connected layer.}
 \end{figure*}

We name the dataset VTechAGP and propose the academic to the general audience text paraphrase task. 
Let $D=\left \{ X_{n},Y_{n},A_{n} \right \}_{n=1}^{N}$ denote a dataset where $X_n$ is the source document, $Y_{n}$ is the target document, $A_{n}$ is the auxiliary information (subject terms, which are the keywords in ETDs provided by the author) of the source document, and $N$ is the number of documents.
Given an input sequence of words $X=[x_1, \cdots, x_{S}]$ with length $S$ and auxiliary information $A=[a_1, \cdots, a_{A}]$ with length $A$, we aim to generate an output sequence of words $Y=[y_1, \cdots, y_T]$ with length $T$ that retains the original meaning as $X$, but reduces the complexity to improve readability and comprehension without domain knowledge.
In this task, our goal is to find the set of model parameters that maximize $\prod _{n=1}^{N}p_{model}(Y_{n}|X_{n})$.

\section{Approach}~\label{sec:approach} 

Figure~\ref{fig:framework} shows our proposed DSPT5 with two main components: dynamic soft prompt generation with hybrid loss and crowd sampling decoding strategy.

\subsection{Backbone}~\label{sec:backbone}
A generative language model prescribes the generation of a sentence as a sequence of word predictions based on context. 
T5~\cite{raffel2020exploring} has shown high performance and few-shot abilities on various language understanding tasks, including text simplification and language translation~\cite{brown2020language}. 
The decoding capacity of the generative language model can help to generate arbitrary content well when given appropriate prompts.
Therefore, we use the pre-trained version of T5 to initialize our model DSPT5.
We fine-tune T5 with dynamic prompts on the training set of VTechAGP~\footnote{FLAN-T5 shows a worse performance based on our task}. 
During the inference stage, we implement the crowd sampling decoding strategy to better select word candidates for text generation.

\subsection{Dynamic Soft Prompt Generation}
In our proposed DSPT5, we introduce an automatic customized soft prompt generation process that includes three main steps: (1) Dynamic Prompt Generation, (2) Soft Prompt Encoder, and (3) Model Training with Hybrid Loss.

\subsubsection{Dynamic Prompt Generation}~\label{sec:dynamic_prompt}
To better control the output generated by T5 for our task, we added a prompt template before the source input (academic abstracts).
Instead of using a static prompt template, such as "Generate another version of the provided document for general audiences", we design a dynamic prompt template that can learn to generate academic keywords to adapt to different academic domains.
As opposed to adding the keywords directly in front of the prompt~\cite{blinova2023simsum}, we designed a template (which is used in the experiments): "Generate the document by reducing the domain knowledge of " + keywords, where the keywords (auxiliary information) $K^{pos}=[k^{pos}_{1}, \cdots, k^{pos}_{N}]$ are generated by KeyBERT~\cite{grootendorst2020keybert}~\footnote{\href{https://github.com/MaartenGr/KeyBERT}{ https://github.com/MaartenGr/KeyBERT}} and sorted by their importance score:
\begin{equation}~\label{equ:0}
K^{pos} = sorted(KeyBERT(X_{i}))
\end{equation}
$N$ is the number of keywords extracted from the source text by KeyBERT.
Since KeyBERT does not support fine-tuning on custom datasets, we develop a soft prompt encoder in Sec.~\ref{sec:soft_prompt} to further fine-tune the embeddings of $K^{pos}$ by minimizing the distance between the embeddings of $K^{pos}$ and the embeddings of the subject terms from the auxiliary information $A$ (golden label). 
Therefore, dynamic prompts can be automatically generated using the keywords extracted from KeyBERT, and the fine-tuned embeddings of such keywords can be obtained by the soft prompt encoder in the inference stage.
Details of loss are introduced in Sec.~\ref{sec:loss}.

\subsubsection{Soft Prompt Encoder}~\label{sec:soft_prompt}
We propose a prompt encoder aiming to generate a soft prompt representing the hidden representation of the text to be generated.
The prompt generator can be any sequence-to-sequence model.
We use the T5~\cite{raffel2020exploring} encoder as the backbone for the prompt encoder.
We first obtain the embeddings $E_{prompt}=[e_1, \cdots, e_{A}]$ and $E_{X}=[e_1, \cdots, e_{S}]$ of a given sequence of tokens $P=[p_1, \cdots, p_{A}]$ and $X=[x_1, \cdots, x_{S}]$ from the partial fixed prompt and the source text, respectively.
Next, we get the output representations of the prompt encoder $r_{i}$, which can be formulated as:
\begin{equation}~\label{equ:1}
\begin{aligned}
h^{key}_{i} &= f_{\phi }(A_{i}) \\
r^{key}_{i} &= ReLU(W\cdot h^{key}_{i}+b)
\end{aligned}
\end{equation}
where $f_{\phi}$ is the prompt (T5) encoder, $h_{i}$ is the last hidden state of the T5 encoder, and $A_{i}$ has the keywords of the subject terms.
A fully connected layer is added after the T5 encoder, where $W \in \mathbb{R}^{n\times d}$ and $b \in \mathbb{R}^{n}$ are trainable. $n$ and $d$ are the length and dimension of the keywords.
Similarly, we get the representations of the extracted keywords $r^{pos}_{i}$:
\begin{equation}~\label{equ:pos}
\begin{aligned}
h^{pos}_{i} &=f_{\phi }(K^{pos}) \\
r^{pos}_{i} &= ReLU(W\cdot h^{pos}_{i}+b)
\end{aligned}
\end{equation}
where $K^{pos}$ is from Equ.~\ref{equ:0}.
We construct negative samples to cover non-academic words. 
As shown in Figure~\ref{fig:framework}, we implement a sampler that randomly samples words from the source text, where the embeddings of the negative samples $r^{neg}_{i}$:
\begin{equation}~\label{equ:neg}
\begin{aligned}
K^{neg} &= Random(X_{i}) \\
h^{neg}_{i} &= f_{\phi }(K^{neg}) \\
r^{neg}_{i} &= ReLU(W\cdot h^{neg}_{i}+b)
\end{aligned}
\end{equation}
where $X_{i}$ is the input source document. The checker in Figure~\ref{fig:framework} ensures that $K^{neg} \cap K^{pos} = \varnothing$.


\subsubsection{Model Training with Hybrid Loss}~\label{sec:loss}
Given the representations of the extracted soft keyword prompts $r^{pos}_{i}$, the embeddings of the prompt template $E_{prompt}$, and the source input $E_{X}$, the output target text generated by our proposed model:
\begin{equation}~\label{equ:model}
\begin{aligned}
P_{\theta}(\cdot | x) &= GLM(\left \{ E_{prompt} \oplus r^{pos} \oplus E_{X}\right \}) \\
\hat{y} &\sim P_{\theta}(\cdot | x)
\end{aligned}
\end{equation}
where $GLM$ is the T5 backbone, $\theta$ is the model parameter, $\oplus$ denotes concatenation, and $\hat{y}$ is the generated general audience document.
DSPT5 is trained to fit the mapping from academic abstracts to general audience abstracts. 
Formally, let the dataset be $D$ with size $N$. 
DSPT5 aims to maximize the standard log-likelihood of the target documents over all training samples of $D$:
\begin{equation}~\label{equ:ce}
\mathcal{L}_{ce} = \sum_{n}^{N}\sum_{s}^{S_n}logP_{\theta }(y_{n,s}|y_{n,< s},x_{n})
\end{equation}
where $y_{n,s}$ is the s-th word of the general audience abstract in the n-th sample, $S_{n}$ is the length of the target output $y_{n}$, and $\theta$ gives the parameters of DSPT5.
To further improve the performance of DSPT5, we add an additional contrastive loss that forces the prompt encoder to generate the embeddings of extracted keywords from KeyBERT $r^{pos}$ that can be more similar to the representations of subject terms $r^{key}$ provided from the source data, while steering away from negative extracted words $r^{neg}$, to encourage the model to bring $r^{key}$ and $r^{neg}$ closer in the learned feature space while pushing dissimilar instances $r^{key}$ and $r^{neg}$ apart.
Therefore, we propose a new hybrid loss function consisting of the cross-entropy loss in Equ.~\ref{equ:ce} and a contrastive loss infoNCE~\cite{oord2018representation} $\mathcal{L}_{nce}$:
\scriptsize
\begin{equation}
\mathcal{L}_{nce} = -log\frac{exp(\frac{r^{key}\cdot r^{pos}}{\tau })}{exp(\frac{r^{key}\cdot r^{pos}}{\tau })+\sum_{r^{neg} \in N^{neg}}exp(\frac{r^{key}\cdot r^{neg}}{\tau })}
\end{equation}
\normalsize
where $r^{key}$ is from Equ.~\ref{equ:1}, $r^{pos}$ is from Equ.~\ref{equ:pos}, $r^{neg}$ is from Equ.~\ref{equ:neg}, $N^{neg}$ is the number of negative samples, and $\tau$ is a temperature hyperparameter.
Details in Appendix Sec.~\ref{sec:explain}.
The final hybrid loss function for DSPT5 model combines $\mathcal{L}_{ce}$ and $\mathcal{L}_{nce}$:
\begin{equation}~\label{equ:loss}
\mathcal{L}_{hybrid} = (1-\lambda)\mathcal{L}_{ce} + \lambda \mathcal{L}_{nce}
\end{equation}
where $\lambda$ is a hyperparameter for $\mathcal{L}_{nce}$.

\subsection{Crowd Sampling Decoding}~\label{sec:decoding}
\begin{figure}[htp] 
 \center{\includegraphics[height=3cm,width=7.75cm]{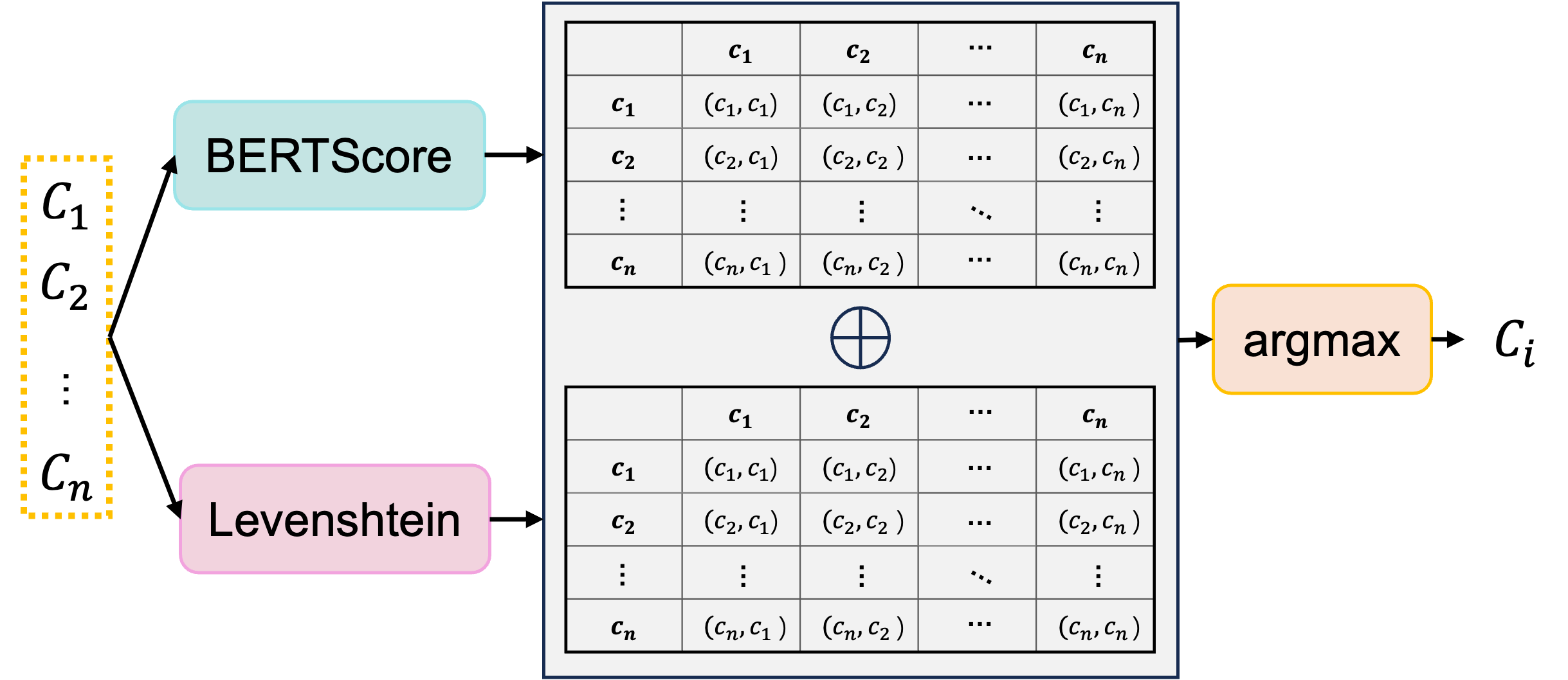}}
 \caption{\label{fig:decoding} The decoding strategy of crowd sampling.}
 \end{figure}

The decoding objective during inference is to select the best candidate $\hat{y}$ among all possible output candidates.
We first implement a stochastic sampling method (temperature sampling~\cite{ackley1985learning}), which selects the next token by sampling from the truncated distribution to generate multiple output candidates. 
Expanding on Equ.~\ref{equ:model}:
\begin{equation}
\tilde{P_{\theta}}(\cdot | x) = \frac{exp(P_{\theta}(\cdot | x)/\tau )}{\sum_{}^{}exp(P_{\theta}(\cdot | x)/\tau )}
\end{equation}
where $\tau \in (0,1]$ is the temperature parameter.
Crowd sampling shows significant performance improvement over standard sampling methods across a wide range of open-ended text generation tasks~\cite{suzgun-etal-2023-follow}.
To select the final generated output $\hat{y}$, we implement crowd sampling as shown in Figure~\ref{fig:decoding} to get the candidate that can maximize the sum of alignments with the whole crowd by comparing each candidate to the other candidates using the alignment functions.
Given a collection of candidate documents $\mathcal{C}$, we get:
\begin{equation}~\label{equ:decoding}
\begin{aligned}
\hat{y} = \underset{c_{i} \in \mathcal{C}}{argmax}(\sum_{y_{i} \in \mathcal{C}}^{}(&BERTScore(c_{i}, y_{i})) \\
                & + \gamma \cdot LEV(c_{i}, y_{i}))
\end{aligned}
\end{equation}
where $\gamma$ is a hyperparameter to adjust the weight for different alignment functions.
To measure both the semantic and structural equivalence of the two texts, we consider BERTScore~\cite{bert-score} and Levenshtein similarity~\cite{Levenshtein1965BinaryCC} as the alignment functions. 
BERTScore computes the cosine similarity between candidates, considering their representations as vectors.
Levenshtein similarity measures the differences between sequences of tokens in the candidates.

\section{Experiments}~\label{sec:experiment} 
\vspace{-6mm}
\subsection{Experimental Setup}
We evaluate the proposed model DSPT5 over eight different colleges in VTechAGP.
The dataset is randomly divided into training and testing sets with a ratio of 0.8:0.2 for all other colleges. 
We discuss the detailed hyperparameters and configurations of DSPT5 in Appendix Sec.~\ref{sec:appendix_setup}.

\subsection{Baselines and Evaluation Metrics}
We compare DSPT5 with the following LLM baselines: Pre-trained LLaMA2~\cite{touvron2023llama}~\footnote{We implement the 7B version of LLaMA2.}, ChatGPT~\cite{brown2020language}~\footnote{\href{https://chatgpt.com/}{https://chatgpt.com/}}, and Claude2~\cite{claude}; Fine-tuned BART~\cite{lewis-etal-2020-bart}, T5~\cite{raffel2020exploring}, and FLAN-T5~\cite{chung2022scaling}. 
Each baseline is fine-tuned on each domain.
Crowd sampling is only added to DSPT5 for evaluation, as it is one component of DSPT5.
All other LLM baselines use their default decoding strategies.
We evaluate DSPT5 with other SOTA LLMs using the following automatic evaluation metrics. 
(1) Embedding-based metrics: BERTScore (F1)~\cite{bert-score}, document-level BLONDE (F1)~\cite{jiang-etal-2022-blonde}, sentence-level and document-level BLEU~\cite{lin-och-2004-orange}; 
(2) Word-based metrics: ROUGE1, ROUGE2~\cite{lin-2004-rouge} and METEOR~\cite{banarjee2005};
(3) End-to-end metrics: COMET~\cite{rei-etal-2022-searching};
(4) Simplicity: SARI~\cite{xu-etal-2016-optimizing}.
(5) Diversity: LCTR (Lexical Translation Consistency Ratio)~\cite{lyu-etal-2021-encouraging}.
(6) Readability: FRES (Flesch Reading-Ease Score)~\cite{flesch1979write};
(7) Toxicity~\cite{nllb2022} for safety evaluation.

\subsection{Results and Analysis}
\subsubsection{Main Results}
\begin{table*}[]
\small
\centering
\tabcolsep=0.09cm
\caption{Performance evaluation of all baselines with twelve different metrics. The abbreviations stand for Agriculture and Life Science, Architecture, Arts and Design, Engineering, Liberal Arts and Human Sciences, Natural Resources and Environment, Science, Business, and Veterinary Medicine.}
\label{tab:main}
\begin{tabular}{lcccccccc|cccccccc}
\hline
Model                & ALS            & AAD            & ENG            & LAHS           & NRE            & SCI            & BUS            & VM             & ALS            & AAD            & ENG            & LAHS           & NRE            & SCI            & BUS            & VM             \\ \hline
                     & \multicolumn{8}{c|}{s-BLEU (\%)}                                                                                                      & \multicolumn{8}{c}{d-BLEU (\%)}                                                                                                       \\ \hline
BART                & 9.25           & 9.59           & 10.67          & 18.01          & \textbf{12.66} & 7.88           & 14.34          & 9.51           & 24.67          & 32.68          & 23.88          & 33.18          & 26.20          & 19.51          & 27.04          & 20.49          \\
T5                 & 7.93           & 11.48          & 8.39           & 16.63          & 11.91          & 7.33           & 15.35          & 8.33           & \textbf{25.61} & 32.79          & 24.82          & 33.58          & 23.80          & 17.00          & 23.77          & 19.56          \\
FLAN-T5            & 8.89           & 12.27          & 8.46           & 17.09          & 6.94           & 7.84           & 14.94          & 8.33           & 18.39          & 28.35          & 17.45          & 31.31          & 17.69          & 15.88          & 30.65          & 18.66          \\
Claude2              & 1.53           & 2.08           & 0.29           & 2.32           & 0.83           & 0.50           & 2.79           & 1.19           & 2.10           & 8.91           & 1.06           & 6.30           & 2..87          & 1.33           & 9.52           & 3.02           \\
ChatGPT            & 4.53           & 5.74           & 4.34           & 6.25           & 4.08           & 4.20  & 8.17           & 4.51           & 15.39          & 15.59          & 13.75          & 16.99          & 14.11          & 12.84          & 17.20          & 13.62          \\
LLaMA2              & 3.39           & 3.22           & 3.56           & 2.39           & 2.77           & 2.93           & 4.02           & 2.39           & 16.70          & 18.51          & 15.02          & 18.64          & 15.51          & 13.21          & 15.66          & 15.54          \\
Ours                 & \textbf{10.84} & \textbf{14.09} & \textbf{11.25} & \textbf{22.12} & 9.19           & \textbf{8.13}  & \textbf{20.30} & \textbf{11.38} & 24.95          & \textbf{33.53} & \textbf{24.98} & \textbf{38.80} & \textbf{28.11} & \textbf{21.31} & \textbf{35.31} & \textbf{23.42} \\ \hline
                     & \multicolumn{8}{c|}{BERTScore (F1\%)}                                                                                                 & \multicolumn{8}{c}{BLONDE (F1 \%)}                                                                                                    \\ \hline
BART              & 85.12          & 82.83          & 84.05          & 86.09          & \textbf{85.58} & 82.74          & 86.59          & 83.94          & 15.04          & \textbf{45.15} & 9.07           & 44.85          & 15.22          & 8.02           & 38.91          & 13.64          \\
T5                  & 85.26          & 83.18          & 84.26          & 85.65          & 84.48          & 81.70          & 85.54          & 83.77          & 14.64          & 38.72          & 8.49           & 43.14          & 33.10          & 7.49           & 34.47          & 25.39          \\
FLAN-T5           & 83.02          & 81.84          & 82.26          & 84.93          & 82.94          & 81.22          & 86.70          & 82.86          & 13.42          & 37.21          & 7.48           & 43.02          & 27.31          & 7.04           & \textbf{39.79} & 27.98          \\
Claude2            & 79.48          & 80.02          & 77.05          & 80.95          & 79.96          & 77.43          & 83.25          & 79.01          & 10.41          & 21.13          & 2.75           & 17.39          & 10.37          & 3.06           & 21.62          & 4.94           \\
ChatGPT            & 84.67          & 82.48          & 83.52          & 84.75          & 84.65          & \textbf{83.05} & 85.17          & 83.89          & 27.73          & 30.99          & \textbf{11.80} & 30.67          & 26.70          & \textbf{11.13} & 11.54          & 21.79          \\
LLaMA2              & 83.58          & 80.97          & 82.46          & 83.66          & 83.31          & 81.46          & 83.14          & 82.74          & 11.38          & 37.39          & 8.25           & 32.77          & 30.56          & 7.63           & 30.01          & 10.82          \\
Ours                 & \textbf{85.48} & \textbf{83.70} & \textbf{84.41} & \textbf{87.27} & 85.51          & 82.90          & \textbf{87.97} & \textbf{84.20} & \textbf{36.75} & 44.11          & 9.62           & \textbf{48.51} & \textbf{36.60} & 8.22           & 35.53          & \textbf{30.69} \\ \hline
                     & \multicolumn{8}{c|}{ROUGE1 (\%)}                                                                                                      & \multicolumn{8}{c}{ROUGE2 (\%)}                                                                                                       \\ \hline
BART                 & \textbf{55.25} & 49.86          & 51.26          & 57.92          & \textbf{57.55} & \textbf{47.49} & 57.78          & \textbf{50.33} & 27.83          & 27.06          & 25.98          & 36.55          & \textbf{31.16} & 20.42          & 33.75          & 22.38          \\
T5                 & 53.94          & 49.74          & 50.88          & 55.34          & 52.07          & 43.40          & 52.22          & 46.66          & \textbf{29.72} & 27.34          & 25.48          & 35.08          & 27.19          & 18.45          & 28.64          & 21.39          \\
FLAN-T5            & 47.44          & 45.27          & 45.22          & 54.07          & 47.37          & 42.48          & 57.71          & 46.77          & 21.54          & 22.17          & 19.20          & 32.30          & 18.93          & 16.87          & 35.75          & 19.63          \\
Claude2               & 28.47          & 36.13          & 21.95          & 36.44          & 33.23          & 23.13          & 44.04          & 29.75          & 5.42           & 9.43           & 2.53           & 9.47           & 6.34           & 3.14           & 14.63          & 5.03           \\
ChatGPT             & 51.50          & 45.61          & 47.97          & 50.32          & 51.38          & 46.06          & 50.35          & 48.31          & 20.08          & 16.75          & 17.88          & 21.89          & 19.67          & 16.55          & 21.76          & 17.54          \\
LLaMA2              & 44.43          & 38.30          & 40.24          & 39.92          & 43.22          & 38.20          & 41.52          & 42.49          & 24.82          & 23.30          & 21.86          & 27.59          & 24.92          & 19.37          & 23.01          & 22.47          \\
Ours                 & 54.48          & \textbf{50.84} & \textbf{51.73} & \textbf{59.67} & 56.47          & 47.14          & \textbf{60.75} & 50.28          & 28.58          & \textbf{27.55} & \textbf{27.02} & \textbf{40.24} & 30.83          & \textbf{21.97} & \textbf{39.61} & \textbf{24.63} \\ \hline
                     & \multicolumn{8}{c|}{METEOR (\%)}                                                                                                      & \multicolumn{8}{c}{COMET (\%)}                                                                                                        \\ \hline
BART                 & 41.27          & 39.57          & 39.28          & 48.65          & \textbf{45.71} & 35.25          & 44.66          & 38.28          & 81.52          & 75.52          & 80.76          & 80.28          & 82.14          & 78.56          & 83.14          & 80.35          \\
T5                  & 40.34          & 40.84          & 39.84          & 46.86          & 40.09          & 31.77          & 40.04          & 33.81          & 80.45          & 74.77          & 80.37          & 76.19          & 75.66          & 70.69          & 77.69          & 73.18          \\
FLAN-T5              & 32.97          & 35.62          & 33.52          & 46.04          & 34.64          & 31.36          & 47.86          & 33.85          & 77.14          & 73.26          & 78.03          & 78.50          & 80.02          & 75.65          & 82.52          & 79.04          \\
Claude2               & 17.50          & 28.53          & 14.96          & 25.81          & 20.88          & 15.34          & 30.85          & 20.59          & 74.31          & 74.38          & 67.74          & 75.53          & 78.02          & 68.72          & 82.70          & 75.05          \\
ChatGPT              & 38.94          & 36.86          & 37.47          & 41.68          & 39.71          & 36.06          & 40.56          & 36.09          & \textbf{85.49} & \textbf{80.30} & \textbf{84.00} & \textbf{83.29} & \textbf{85.79} & \textbf{82.98} & \textbf{85.29} & \textbf{84.10} \\
LLaMA2            & \textbf{42.53} & 39.88          & \textbf{41.13} & 43.68          & 44.10          & \textbf{39.51} & 41.25          & \textbf{41.47} & 82.11          & 77.85          & 82.22          & 81.17          & 82.65          & 79.90          & 82.88          & 80.84          \\
Ours                 & 40.50          & \textbf{42.51} & 40.74          & \textbf{52.67} & 45.05          & 36.65          & \textbf{50.56} & 38.01          & 81.37          & 75.90          & 80.64          & 81.76          & 82.09          & 77.15          & 84.07          & 78.36          \\ \hline
\multicolumn{1}{c}{} & \multicolumn{8}{c|}{SARI}                                                                                                             & \multicolumn{8}{c}{LTCR (\%)}                                                                                                         \\ \hline
BART                & 37.22          & 34.08          & 36.87          & 36.34          & 37.36          & 36.91          & 42.43          & 36.45          & 59.36          & 60.85          & 58.29          & 55.53          & 59.97          & 63.83          & 61.98          & 64.02          \\
T5                 & 37.89          & 35.89          & 37.88          & 36.83          & 36.11          & 37.21          & 47.75          & 36.92          & 57.61          & 60.82          & 56.18          & 52.27          & 60.51          & 62.74          & 58.84          & 66.01          \\
FLAN-T5             & 36.73          & 34.12          & 36.78          & 35.43          & 34.83          & 37.76          & 48.62          & 36.71          & 61.41          & 63.91          & 60.63          & 56.89          & 64.59          & \textbf{84.12} & 61.19          & 67.16          \\
Claude2             & 29.34          & 32.44          & 28.85          & 28.48          & 30.17          & 31.19          & 34.79          & 32.08          & \textbf{71.66} & \textbf{72.94} & \textbf{68.61} & \textbf{68.46} & \textbf{72.21} & 73.19          & \textbf{70.99} & \textbf{77.28} \\
ChatGPT            & 37.83          & 35.41          & 36.90          & 34.50          & 36.61          & 38.53          & 40.07          & 38.07          & 58.87          & 65.38          & 57.61          & 59.64          & 61.14          & 61.97          & 64.28          & 63.35          \\
LLaMA2              & \textbf{46.43} & \textbf{40.51} & \textbf{43.76} & \textbf{47.45} & \textbf{46.31} & \textbf{43.53} & 44.99          & \textbf{44.55} & 57.90          & 61.00          & 57.90          & 55.70          & 61.10          & 62.56          & 64.37          & 62.78          \\
Ours                 & 37.31          & 36.01          & 38.21          & 38.95          & 37.23          & 37.11          & \textbf{48.67} & 36.50          & 56.81          & 61.33          & 55.25          & 51.25          & 58.35          & 62.18          & 56.46          & 63.46          \\ \hline
                     & \multicolumn{8}{c|}{FRES}                                                                                                             & \multicolumn{8}{c}{Mean Toxicity (\%)}                                                                                                \\ \hline
BART               & 32.22          & 40.18          & 30.60          & 27.66          & 39.47          & 30.80          & 19.91          & 29.79          & 0.06           & 0.83           & 0.07           & 0.71           & 0.04           & 0.08           & 0.09           & 0.06           \\
T5                  & 32.83          & 41.60          & 31.01          & 27.96          & 40.69          & 31.62          & 20.72          & 31.82          & 0.12           & 0.74           & 0.08           & 1.00           & 0.04           & 0.11           & 0.10           & 0.06           \\
FLAN-T5           & 40.89          & 40.89          & 31.41          & 29.48          & 40.99          & 32.63          & 22.34          & 31.62          & 0.07           & 0.54           & 0.08           & 1.40           & 0.04           & 0.11           & 0.08           & 0.05           \\
Claude2             & 31.99          & 41.97          & 13.65          & 21.60          & 40.45          & 12.63          & 26.40          & 29.45          & 0.06           & 0.49           & 0.10           & 0.93           & 0.09           & 0.11           & 0.10           & 0.23           \\
ChatGPT             & \textbf{42.11} & \textbf{42.41} & \textbf{33.85} & \textbf{31.72} & \textbf{41.70} & \textbf{34.15} & \textbf{29.69} & \textbf{42.21} & \textbf{0.03}  & \textbf{0.06}  & 0.05           & \textbf{0.32}  & 0.03           & \textbf{0.05}  & 0.09           & \textbf{0.03}  \\
LLaMA2             & 34.26          & 41.60          & 31.92          & 30.70          & 33.54          & 32.73          & 22.85          & 32.53          & 0.11           & 0.55           & 0.16           & 1.06           & 0.08           & 0.13           & 0.13           & 0.11           \\
Ours                 & 33.65          & 41.50          & 31.62          & 28.17          & 41.19          & 32.33          & 21.74          & 33.34          & 0.12           & 1.19           & \textbf{0.04}  & 0.97           & \textbf{0.02}  & 0.06           & \textbf{0.04}  & \textbf{0.03}  \\ \hline
\end{tabular}
\end{table*}

\begin{table*}[]
\small
\centering
\tabcolsep=0.04cm
\caption{Ablation study over DSPT5 components on VTechAGP across eight colleges. DSPT5\textsubscript{dynamic} is the model only using a dynamic hard prompt from keyBERT. DSPT5\textsubscript{dyn+soft} is the model using a dynamic soft prompt from the prompt encoder. DSPT5\textsubscript{dyn+soft+con} is the model fine-tuned using dynamic soft prompt with contrastive learning loss. DSPT5\textsubscript{all} is the final version of the model using dynamic soft prompt with contrastive learning for training and crowd sampling for inference.}
\label{tab:ablation}
\begin{tabular}{lcccccccccccc}
\hline
Model          & s-BLEU & d-BLEU & BERTScore & BLONDE & ROUGE1 & ROUGE2 & METEOR & COMET & SARI & LTCR & FRES & Toxicity \\ \hline
T5\textsubscript{pre-trained}             & 3.76                           & 0.84                            & 78.61                           & 11.34                           & 23.74                           & 12.06                           & 11.74                           & 59.66                           & 29.81 & \textbf{78.01} & \textbf{36.36} & 0.23          \\
T5\textsubscript{fine-tuned}            & 6.02                           & 19.14                           & 82.14                           & 22.14                           & 45.53                           & 22.14                           & 36.44                           & 68.74                           & 36.55 & 62.77 & 31.54 & 0.20       \\
Ours\textsubscript{dynamic} & 8.78                           & 22.83                           & 83.83                           & 28.43                           & 49.77                           & 25.33                           & 38.35                           & 77.28                           & 36.71 & 60.19 & 32.31 & 0.24          \\
Ours\textsubscript{dyn+soft}   & 9.87                           & 25.79                           & 84.43                           & 34.92                           & 51.14                           & 26.98                           & 40.64                           & 78.91                           & 37.03 & 59.11 & 32.34 & \textbf{0.11}      \\
Ours\textsubscript{dyn+soft+con}   & 9.88                           & 26.78                           & \textbf{84.58}                           & 33.99                           & \textbf{51.66}                           & 26.82                           & 40.90                           & 78.54                           & 37.03 & 59.05 & 32.33 & 0.13      \\ \hline
Ours\textsubscript{all}          & \textbf{9.93} & \textbf{26.99} & 84.52 & \textbf{35.75} & 51.45 & \textbf{27.59} & \textbf{41.40} & \textbf{79.06} & \textbf{37.05} & 58.77 & 32.39                      & 0.17      \\ \hline
\end{tabular}
\end{table*}

The experiment results on VTechAGP are shown in Table~\ref{tab:main}.
We observe that: First, for the accuracy measurement between the generated text and the ground truth (i.e., embedding-based metrics and word-based metrics), our proposed model achieves the best performance in most cases. In addition, fine-tuned models (i.e. Bart, T5, FLAN-T5) always show better translation performance (BLEU, BERTScore, and ROUGE) than pre-trained LLMs (i.e. Claude2, ChatGPTP, LLaMA2). We believe this is because when the reference text (ground truth) is used to update the model's parameters during the fine-tuning process, the model adapts to the nuances and characteristics of our specific academic to general audience text paraphrasing task.
In contrast, the pre-trained LLMs are trained on large and diverse datasets and are not optimized for the academic domain.
This indicates that it is possible to fine-tune a lightweight language model that can achieve competitive translation performance with LLMs on the high-quality VTechAGP dataset.

Second, pre-trained LLMs outperform fine-tuned lightweight models in end-to-end metrics, simplicity, diversity, and readability. However, different LLMs show distinctive strengths and capabilities. For example, LLaMA2 performs well at text simplification. Claude2 is only good at generating diverse text, and ChatGPT demonstrates the best performance for end-to-end metrics and readability. 
Each LLM has a unique architecture and has been pre-trained on different datasets, which may explain the observed differences.
Finding an LLM that performs well in all evaluation metrics across all domains is challenging, indicating that there is still much room for improvement in LLMs for academic-to-general-audience text paraphrasing tasks.
Besides, we need to further explore new evaluation metrics that can more accurately reflect the quality of the generated general audience text.

\subsubsection{Ablation Study}
To assess the performance of each component in DSPT5, we conducted the ablation study.
For each run, we randomly sampled 15 data points from each college to construct a balanced dataset for the ablation study (see ablation study setup in Appendix~\ref{sec:appendix_setup}). 
The results are presented in Table~\ref{tab:ablation} as the mean value of five different subsets.
We observe that: 
(1)  Fine-tuning the LLM can significantly improve its performance. This is because fine-tuning the model on a customized dataset can enable it to learn domain-specific nuances and identify task-specific patterns. 
(2) Adding the dynamic soft prompt consisting of domain-specific keywords can further improve the performance. The results verify that adding the extracted academic keywords with the phrase "reduce the domain knowledge of" in the prompt template while training with contrastive-generative loss can reduce the domain knowledge. Besides, the crowd sampling decoding is only implemented in the inference stage, it is worth adding the module for further performance improvement;
(3) The diversity LTCR and readability FRES decrease after fine-tuning T5. We suspect that this is because T5 is pre-trained on diverse and large datasets, resulting in the generated text being represented by a diverse vocabulary. Moreover, the output of T5\textsubscript{pre-trained} is always shortened compared to T5\textsubscript{fine-tuned}, making the output easier to understand (higher FRES). Thus, new evaluation metrics for academic-to-general-audience text paraphrasing tasks are needed in the future.

\subsubsection{Human Evaluation}~\label{sec:human}
\begin{table}[]
\small
\centering
\tabcolsep=0.18cm
\caption{Mean human evaluation ratings 1-5 (the higher the better) of different models on VTechAGP. Reported from left to right are: comprehensiveness, layness, meaning preservation, conciseness, and fluency. ICC stands for the intraclass correlation coefficient.}
\label{tab:human}
\begin{tabular}{lcccccc}
\hline
        & COM & LAY       & MP & CON   & FLU  & ICC     \\ \hline
BART    & 3.83              & 2.83          & 3.67                 & 3.58          & 3.83    &\textbf{0.85}      \\
T5      & 3.61              & 2.95          & 3.35                 & 3.5           & 3.51    &0.60      \\
FlanT5  & 3.71              & 3.13          & 3.41                 & 3.66          & 3.76    &0.38      \\
Claude2 & 3.08              & 2.95          & 2.72                 & 3.55          & 3.26    & 0.67      \\
ChatGPT & \textbf{4.17}     & \textbf{3.66} & 3.83                 & 3.71          & \textbf{4.31} &\textbf{0.85} \\
LLaMA2  & 4.05              & 3.61          & 3.73                 & 3.80           & \textbf{4.31} & 0.39\\
Ours    & \textbf{4.17}     & 3.57          & \textbf{3.90}         & \textbf{3.87} & 3.95     &0.78     \\ \hline
\end{tabular}
\end{table}

Following a similar setting as~\cite{liu-etal-2024-sumsurvey, li-etal-2024-side, song-etal-2024-finesure, kew-etal-2023-bless, devaraj-etal-2021-paragraph}, our evaluation uses a random sample of 20 abstracts from the test split VTechAGP considering the workload. Judges are presented with both the academic abstract and generated general-audience abstracts from seven models for each data sample in a total of 140 abstracts. Using a 1-5 Likert scale, the judges are asked to rate the model output based on five criteria: comprehensiveness, layness, meaning preservation, conciseness, and fluency. 
More details are discussed in detail in Sec.~\ref{sec:appendix_human} in the Appendix.

Human evaluation results are shown in Table~\ref{tab:human}. 
From Table~\ref{tab:human}, we observe that generally ChatGPT, and our proposed DSPT5 show better performance than other baselines.
ChatGPT performs well in comprehensiveness, layness, and fluency, whereas DSPT5 outperforms all other baselines in comprehensiveness, meaning perservation, and conciseness.
Based on the observations, new information with simpler words is always introduced in the generated outputs from ChatGPT.
Those sentences with simpler words make the generated abstracts easier to understand, resulting in better layness.
However, it also significantly changes the meaning of the original sentences and makes the sentences longer with redundancy.
That's why ChatGPT shows the best performance in layness, while not the best in meaning preservation and conciseness.
DSPT5 attempts to paraphrase the abstracts with non-technical words while retaining the original meaning as closely as possible. While keeping the content concise, it may sacrifice fluency between sentences.

For intraclass correlation coefficient (ICC)~\footnote{We use Pingouin package to calculate ICC: \href{https://pingouin-stats.org/build/html/index.html}{https://pingouin-stats.org/build/html/index.html}}, which is used to determine if items or subjects can be rated reliably by different raters, BART, ChatGPT and DSPT5 show good reliability. T5 and Claude2 show moderate reliability. FlanT5 and LLaMA2 show poor reliability according to~\cite{koo2016guideline}. 
Although ChatGPT shows competitive results, it is NOT an open-sourced model, which may raise concerns about data security and high-cost issues.
Alternatively, the DSPT5 can be fully controlled and owned with only a single V100 GPU for fine-tuning jobs for the academic-to-general-audience text paraphrasing task.

\section{Conclusion}
We created VTechAGP~\footnote{Benchmarks: \href{https://github.com/SIGSEGV-0x7/VTechAGP-Benchmark}{https://github.com/SIGSEGV-0x7/VTechAGP-Benchmark}}, an academic-to-general-audience document-level translation benchmark dataset, which consists of academic and general audience abstract pairs with their corresponding auxiliary information such as title, subject terms, etc.
We explore several SOTA LLMs to establish the baseline performance on the text paraphrasing task.
We propose a new model, DSPT5, which includes dynamic soft prompt generation with hybrid loss during the training phase and a new crowd-sampling strategy in the inference stage.
We evaluate the datasets and models from the perspective of document-level embedding-based, word-based, end-to-end metrics, simplicity, diversity, readability, toxicity, and human evaluation.
Extensive experimental results on 8 different colleges of VTechAGP show that DSPT5 achieves comparable results with other SOTA benchmark models. 

\section{Limitations}
We identify the following limitations (remaining challenges) and future directions:

(1) Dataset: Although VTechAGP is a recent and distinctive dataset, it suffers in terms of size. This is due to the fact that the policy of including a general audience version of the abstract for all ETDs submitted to VTechWorks began to be implemented in 2016. 
However, this aspect of the limitation should be gradually overcome as VTechWorks continues to collect and maintain this collection of documents. With the increase in graduate student enrollment at Virginia Tech in recent years, it is safe to assume that VTechAGP will continue to grow in the coming years.
In addition, VTechAGP consists of pairs of abstracts carefully written by students, encapsulating their work during their time in graduate school. With about 50\% of the corpus composed of Ph.D. dissertations and the other half composed of M.S. theses, we believe that VTechAGP should achieve good quality due to the amount of time and effort put into the creation process. As the size of the dataset grows, VTechAGP will become more robust for text paraphrasing tasks in the near future.

(2) Model:
While transformer-based LLMs retain their popularity for text generation tasks, our experiment results in Sec.~\ref{sec:experiment} show that there is still a huge room for further improvement of LLMs in academic to general audience text paraphrasing tasks. 
With the rapid development of LLMs, more robust and efficient LLMs are emerging. For example, the recently released model Gemini~\cite{team2023gemini} shows great potential and competitive results compared to ChatGPT and LLaMA2 in natural language understanding-related tasks.
In addition, some recently released lightweight decoder-only models (e.g., \href{https://www.microsoft.com/en-us/research/blog/phi-2-the-surprising-power-of-small-language-models/}{Phi-2}~\cite{li2023textbooks}) also exhibit great potential for text generation tasks.
As the VTechAGP dataset grows, it will be able to support fine-tuning for more powerful open-source LLMs.

(3) Evaluation Metric: 
Although we have implemented twelve different automatic evaluation metrics, including document-level embedding-based metrics, word-based metrics, end-to-end metrics, simplicity, diversity, readability, and toxicity, we still lack a representative metric that can evaluate the quality of the generated text in terms of general audience understanding while remaining scientifically accurate and easy to comprehend. 
The existing evaluation metrics for simplicity such as SARI~\cite{xu-etal-2016-optimizing} focus on measuring the goodness of words that are added, deleted, and kept by the system.
In addition, regarding the formula of FRES~\cite{flesch1979write}, such readability evaluation metric only considers the number of words and the number of sentences when evaluating the ease of understanding for children of different grades.
Instead of just simplifying sentence structure and using words with fewer letters, our task aims to reduce the domain knowledge of the generated text while remaining scientifically correct, so that the general audience or people/researchers in other research fields (departments) can still understand the core idea for further interdisciplinary collaboration.
New automatic evaluation metrics or large-scale human evaluation approaches for academic-to-general-audience text paraphrasing tasks need to be explored in future work.

\section*{Acknowledgments}
This project was made possible in part by the Institute of Museum and Library Services (LG-256638-OLS-24).

\bibliography{custom}

\appendix

\section{Dataset Comparison}~\label{sec:appendix}
In this section, we explore different features (i.e. document-level or sentence-level, multiple domains or single domain, with auxiliary information or not, in English or other languages) of existing public datasets for text simplification and text paraphrasing, which motivated us to collect and publish a new academic-to-general-audience text paraphrasing dataset VTechAGP, which is a document-level, multi-domain English text paraphrasing dataset with domain-specific auxiliary information.
We show the comparison of our proposed dataset VTechAGP with other existing public datasets. The details of features are displayed in Table~\ref{tab:example}.
\begin{table*}[htp]
\caption{Comparison between VTechAGP and other text simplification or paraphrase datasets. Columns indicate whether the dataset has the feature.}
\label{tab:example}
\centering
\begin{tabular}{l|c|c|c|c}
\hline
               & Doc. & Multi Domains & Side Info. & Eng. \\ \hline
Newsela~\cite{xu-etal-2015-problems}        &     \cmark      &      \xmark         &     \xmark       &    \cmark     \\
WikiSmall~\cite{zhu-etal-2010-monolingual}      &      \xmark       &   \xmark             &      \xmark       &   \cmark      \\
WikiLarge~\cite{zhang-lapata-2017-sentence}      &     \xmark        &   \xmark             &         \xmark    &  \cmark       \\
WikiAuto~\cite{acl/JiangMLZX20}       &    \cmark        &     \xmark           &       \cmark     &      \cmark   \\
TurkCorpus~\cite{xu2016optimizing}     &   \xmark          &    \xmark            &        \xmark     &   \cmark      \\
ASSET~\cite{alva-manchego-etal-2020-asset}          &   \xmark          &   \xmark             &       \xmark      &    \cmark     \\
MedLane~\cite{luo2022benchmarking}        &    \xmark         &      \xmark          &       \xmark      &  \cmark       \\
OneStopEnglish~\cite{vajjala2018onestopenglish} &     \xmark        &   \xmark             &      \xmark       &    \cmark     \\
Lexica~\cite{hewett-stede-2021-automatically}         &    \cmark        &   \xmark             &       \cmark     &  \xmark        \\
DEplain~\cite{stodden-etal-2023-deplain}        &   \cmark         &         \cmark      &        \cmark   &   \xmark       \\
Quora~\cite{ijcai2017p579}          &     \xmark        &      \xmark          &        \xmark     &   \cmark      \\
Paralex~\cite{fader2013paraphrase}        &       \xmark      &    \xmark            &       \xmark      &   \cmark      \\
MRPC~\cite{dolan-brockett-2005-automatically}           &    \xmark         &  \xmark              &       \xmark      &    \cmark     \\
STS~\cite{cer-etal-2017-semeval}            &   \xmark          &    \xmark            &     \xmark        &    \cmark     \\
SNLI~\cite{bowman2015large}           &   \xmark          &   \xmark             &       \cmark     &    \cmark     \\
ParaNMT-50M~\cite{wieting2017paranmt}        &   \xmark          &       \xmark         &       \xmark      & \cmark        \\
MSCOCO~\cite{lin2014microsoft}         &     \xmark        &     \xmark           &      \cmark      &    \cmark     \\ \hline
\textbf{VTechAGP (ours) }     &       \cmark     &  \cmark             &   \cmark         &    \cmark     \\ \hline
\end{tabular}
\end{table*}

\begin{table}[htp]
    \centering
    \small
    \caption{Description of CSV File Columns in the Dataset}
    \label{tab:dataset_columns}
    \begin{tabular}{lp{5.2cm}}
        \toprule
        Column Name & Description \\
        \midrule
        identifier\_uri & Persistent identifier (CNRI handle) \\
        title & Title \\
        abstract & Regular abstract \\
        abstract\_general & General audience abstract \\
        subject\_terms & List of subject terms  \\
        discipline & Field of study for the degree awarded \\
        department & Name of the academic department \\
        degree & Degree awarded \\
        degree\_level & Degree level (`doctoral' or `masters') \\
        type & Type of ETD (`thesis' or `dissertation') \\
        \bottomrule
    \end{tabular}
\end{table}

\begin{table*}[h]
\centering
\caption{An example of academic abstract and general-audience abstract pair.}
\label{tab:pair}
\begin{tabular}{|p{15cm}|}
\hline
\textbf{Academic Abstract} \\
\hline
 Breath-first search (BFS) is a fundamental building block in many graph-based applications, but it is difficult to optimize for a field-programmable gate array (FPGA) due to its irregular memory-access patterns. Prior work, based on hardware description languages (HDLs) and high-level synthesis (HLS), addresses the memory-access bottleneck of BFS by using techniques such as data alignment and compute-unit replication on FPGAs. The efficacy of such optimizations depends on factors such as the sparsity of target graph datasets. Optimizations intended for sparse graphs may not work as effectively for dense graphs on an FPGA and vice versa. This thesis presents two sets of FPGA optimization strategies for BFS, one for near-hypersparse graphs and the other designed for sparse to moderately dense graphs. For near-hypersparse graphs, a queue-based kernel with maximal use of local memory on FPGA is implemented. For denser graphs, an array-based kernel with compute-unit replication is implemented. Across a diverse collection of graphs, our OpenCL optimization strategies for near-hypersparse graphs delivers a 5.7x to 22.3x speedup over a state-of-the-art OpenCL implementation, when evaluated on an Intel Stratix~10 FPGA. The optimization strategies for sparse to moderately dense graphs deliver 1.1x to 2.3x speedup over a state-of-the-art OpenCL implementation on the same FPGA. Finally, this work uses graph metrics such as average degree and Gini coefficient to observe the impact of graph properties on the performance of the proposed optimization strategies.      \\
\hline
\textbf{General-Audience Abstract} \\
\hline
A graph is a data structure that typically consists of two sets -- a set of vertices and a set of edges representing connections between the vertices. Graphs are used in a broad set of application domains such as the testing and verification of digital circuits, data mining of social networks, and analysis of road networks. In such application areas, breadth-first search (BFS) is a fundamental building block. BFS is used to identify the minimum number of edges needed to be traversed from a source vertex to one or many destination vertices. In recent years, several attempts have been made to optimize the performance of BFS on reconfigurable architectures such as field-programmable gate arrays (FPGAs). However, the optimization strategies for BFS are not necessarily applicable to all types of graphs. Moreover, the efficacy of such optimizations oftentimes depends on the sparsity of input graphs. To that end, this work presents optimization strategies for graphs with varying levels of sparsity. Furthermore, this work shows that by tailoring the BFS design based on the sparsity of the input graph, significant performance improvements are obtained over the state-of-the-art BFS implementations on an FPGA.\\
\hline
\end{tabular}

\end{table*}

\section{Dataset Analysis}~\label{sec:appendix_analysis}
The original dataset does not include the college information. We manually assign each text sample to the college subset based on the department information on each official college website from Virginia Tech. Such data preprocessing steps introduce a few duplicates in the data because: (1) Some departments are involved in more than one college. For example, according to Virginia Tech's official website, the Department of Biology Systems Engineering appears both in the College of Engineering~\footnote{\href{https://eng.vt.edu/academics/departments.html}{https://eng.vt.edu/academics/departments.html}} and the College of Agriculture and Life Sciences~\footnote{\href{https://www.cals.vt.edu/departments-and-school.html}{https://www.cals.vt.edu/departments-and-school.html}}. (2) According to VT each College's website, some departments (i.e. Environmental Science and Engineering, Counselor Education, etc.) can not be assigned to a specific college. We use fuzzy matches to find similar but not exact matches to assign data to colleges. Note that such a processing step may introduce a small number of duplicate files. 

After processing, VTechAGP contains a total of 4,938 document pairs (academic abstract and general audience abstract), where 52.8\% of the documents are from Ph.D. dissertations and 47.2\% of the documents are from Master's theses.
Table~\ref{tab:dataset} shows the data statistics of VTechAGP across eight different colleges. 
VTechAGP is divided into eight colleges because data from different colleges exhibits distinct statistics. 
In addition, the dataset divided into colleges can also support potential cross-domain research tasks and provide convenience for researchers focusing on a specific domain in the future. 
The document pairs are categorized based on colleges since authors from the same college may share some common terminology and background knowledge. Therefore, these categories should exhibit similar statistics.
Table~\ref{tab:dataset} reports some basic dataset statistics, including the number of documents for each college, the average number of sentences for each document, and the average sentence length for the source and target texts, respectively.
We also report textual and lexical diversity (MTLD), which reflects the average number of words in a row for which a certain type-token ratio is maintained~\cite{mccarthy2005assessment}.
To better illustrate the difference in readability between the source and target texts, we report Readability Consensus,\footnote{\href{https://pypi.org/project/textstat/}{https://pypi.org/project/textstat/}} which is a combined evaluation of Flesch Kincaid Grade, Flesch Reading Ease, SMOG Index, Coleman-Liau Index, Automated Readability Index, Dale-Chall Readability Score, Linsear Write Formula, and Gunning FOG Formula.
From Table~\ref{tab:dataset}, we observe that the general audience abstracts have fewer sentences and shorter sentence lengths compared to the academic abstracts.
Furthermore, academic abstracts have better lexical diversity than general audience abstracts, except for the College of Architecture, Arts and Design, and the College of Business. 
In general, academic abstracts have a higher level of comprehension difficulty compared to general audience abstracts, except the College of Engineering.
Analysis for the College of Engineering, which represents almost half of the total dataset, is discussed below.

\begin{table}[htp]
\small
\caption{Dataset statistics of VTechAGP over 13 departments in the College of Engineering. Statistics are reported in the format of academic (source) / general audience (target) doc.}
\label{tab:dataset_department}
\centering
\begin{tabular}{lc}
\hline
Department                  & Readability Consensus \\ \hline
Aerospace Eng.              & 14th-15th / 14th-15th \\
Biomedical Eng.             & 13th-14th / 10th-11th \\
Building Construction       & 15th-16th / 15th-16th \\
Chemical Eng.               & 16th-17th / 11th-12th \\
Civil Eng.                  & 16th-17th / 15th-16th \\
Computer Sci.               & 14th-15th / 15th-16th \\
Electrical \& Computer Eng. & 15th-16th / 15th-16th \\
Eng. Education              & 17th-18th / 16th-17th \\
Environmental Sci. \& Eng.  & 15th-16th / 14th-15th \\
Industrial \& Systems Eng.  & 16th-17th / 15th-16th \\
Materials Sci. \& Eng.      & 16th-17th / 11th-12th \\
Mechanical Eng.             & 15th-16th / 14th-15th \\
Mining Eng.                 & 15th-16th / 15th-16th \\ \hline
\end{tabular}
\end{table}
The distinct category (College of Engineering) contains the largest number of document pairs, representing almost half of the total dataset, due to the large number of students from 13 different departments in the College of Engineering. To better understand the characteristics of this category, we divided the documents in the College of Engineering based on their corresponding departments. Details are given in Table~\ref{tab:dataset_department}. Thus, we expect document pairs within this college to be more diverse compared to others. 
From Table~\ref{tab:dataset_department}, we observe that documents from most departments in the College of Engineering have a similar readability consensus between academic abstracts and general audience abstracts. 
The only outlier is the Department of Computer Science, which shows that general audience abstracts are more difficult to understand than academic abstracts.
On the other hand, the Department of Electrical and Computer Engineering alone represents 15.2\% of the total dataset, while maintaining an above-average readability consensus for its general audience abstracts, thus contributing to the higher readability consensus for the College of Engineering.
We conjecture as follows: 1) Various complex terminologies from some research fields are inevitably retained in general audience abstracts because they cannot be replaced by simple terms. 2) Concepts in certain areas of engineering research are difficult to explain or rewrite into short and concise sentences. 3) Some engineering students are not as good at simplification writing as students from other colleges.

\section{Parameter Settings}~\label{sec:appendix_setup}
For the hyperparameters and configuration of DSPT5, we implement DSPT5 in PyTorch and optimize it with the AdamW optimizer.
For pre-trained LLMs, we retrieve the generated output through their API by giving the prompt "Generate another version of the provided document for the general audience".
We use grid search to tune the hyperparameters. The learning rate is $\in \left \{ 5e-2, 5e-3, 5e-4, 5e-5 \right \}$.
The contrastive loss weight $\lambda$ in Equ.~\ref{equ:loss} and the alignment function weight $\gamma$ in Equ.~\ref{equ:decoding} are chosen from 0.1 to 0.9 with a step size of 0.2. 
The number of candidates $C_{n}$ in Sec.~\ref{sec:decoding} is chosen from $\left \{ 4, 8, 16 \right \}$.
The batch size is 4.
The length of the source text, fixed prompt template, and keywords are 512, 16, and 16, respectively.
The fixed prompt template used for baselines is: `Generate another version of the provided document for general audiences.'.
We choose the hyperparameters based on the validation set. Because the dataset is not a balanced dataset, where College of Engineering takes up most of the datasets. We split a validation set in the domain of College of Engineering documents. This is because as shown in Table~\ref{tab:dataset}, it is not realistic to have a validation set in the Business domain, which has only 63 documents in total. 
The size of the dataset will only grow as new master's theses or PhD dissertations are submitted. In the future, when the dataset is large enough, we will split out a validation set that includes data from all colleges.
The final learning rate we used is 5e-5. However, this can be changed according to different batch sizes when fine-tuning the model. The parameters we used are searched by a validation set from the engineering college documents (which takes the highest percentage of the total documents). The contrastive loss weight is 0.3, the alignment function weight is 0.1, and the number of candidates is 16. If people want to work in a different domain (other than the College of Engineering), different parameters may have different performances. For the temperature in contrastive loss, we use the InfoNCE loss from~\cite{oord2018representation}, and we use the default temperature value of 0.1. The temperature in decoding is 0.5.
For model details shown in Figure~\ref{fig:framework}, we use the same T5-encoder that shares parameters to keep our model DSPT5 as small (efficient) as possible. 
Here are the exact versions for LLM used: ChatGPT: gpt-3.5-turbo-0613 Claude2: Claude2.1 LLaMA: LLaMA2-7b BART: bart-large-cnn FlanT5: flan-t5-base T5: t5-base.

For the ablation study setup, the model is trained on the entire training dataset without sampling. We only do the sampling on the evaluation dataset for the ablation study. Since the data split is 0.8:0.2 on each college domain, the evaluation set is also not a balanced dataset. 
In total, there are about 988 documents in the evaluation set, which is used for the evaluation in Table~\ref{tab:main}.
After sampling, we have around 120 documents (which is now a balanced evaluation set covering all colleges) for the evaluation set ONLY used for the ablation study Table~\ref{tab:ablation}.
We reported the average performance including all colleges in Table~\ref{tab:ablation} for the ablation study.
Since the College of Engineering takes up almost half of the data shown in Table~\ref{tab:dataset}, the result of the average performance will be heavily influenced by the performance of the College of Engineering. 
Therefore, we sampled a balanced evaluation set to report the ablation study.

\section{More Results}~\label{sec:more}
In this section, we provide the combined test set results of all eight colleges, which is shown in Tabel~\ref{tab:all}.
From Table~\ref{tab:all}, we observe that our proposed DSPT5 outperforms all other baselines in the evaluation metric of s-BLEU, d-BLEU, BERTScore, BLONDE, ROUGE and METEOR. The closed-source model ChatGPT shows better performance in system-level evaluation metrics such as COMET. We conjecture that this is probably because such pre-trained LLMs (i.e. ChatGPT, Claude2, LLaMA2) are pre-trained on large quantities of documents, which makes these models generate text more fluently, resulting in better system-level evaluation metrics. Alternatively, our proposed model DSPT5 is fine-tuned on our customized dataset VTechAGP, which results in better performance in word-level evaluation metrics.
\begin{table*}[]
\small
\caption{Combined test results of all eight colleges.}
\label{tab:all}
\tabcolsep=0.06cm
\begin{tabular}{lcccccccccccc}
\hline
        & s-BLEU         & d-CLEU         & BERTScore      & BLONDE         & ROUGE1         & ROUGE2         & METEOR         & COMET          & SARI           & LTCR           & FRES           & Toxicity      \\ \hline
BART    & 11.24          & 25.09          & 84.28          & 16.82          & 52.33          & 27.15          & 40.58          & 80.27          & 36.79          & 59.19          & 31.12          & 0.19          \\
T5      & 9.75           & 25.16          & 84.15          & 17.20          & 50.63          & 26.30          & 39.62          & 77.71          & 37.56          & 57.44          & 31.73          & 0.24          \\
FLAN-T5 & 9.77           & 19.93          & 82.64          & 16.26          & 46.57          & 21.23          & 35.24          & 77.66          & 36.65          & 63.99          & 32.98          & 0.28          \\
Claude2 & 0.88           & 2.50           & 78.28          & 7.03           & 26.44          & 4.57           & 18.04          & 70.78          & 29.62          & \textbf{70.16} & 19.57          & 0.23          \\
ChatGPT & 4.70           & 14.36          & 83.77          & 17.65          & 48.45          & 18.52          & 38.07          & \textbf{83.84} & 36.86          & 59.40          & \textbf{35.33} & \textbf{0.08} \\
LLaMA2  & 3.21           & 15.71          & 82.58          & 14.70          & 40.43          & 22.83          & 41.49          & 81.54          & \textbf{44.51} & 58.78          & 32.50          & 0.29          \\
Ours    & \textbf{12.41} & \textbf{27.02} & \textbf{84.75} & \textbf{21.11} & \textbf{52.71} & \textbf{28.59} & \textbf{42.15} & 80.21          & 38.01          & 56.50          & 32.32          & 0.23          \\ \hline
\end{tabular}
\end{table*}

\section{Case Study}~\label{sec:case}
Table~\ref{tab:case} shows an example of general audience abstracts generated by different LLMs on VTechAGP (Department of Chemical Engineering).
The first few sentences of the document are displayed.
Fine-tuned T5 and FLAN-T5 show similar outputs with source text, so we have not included them in Table~\ref{tab:case}.
The key idea of the source text is: Polymers are important in life and its manufacturing process is critical to industry.
We observe that the pre-trained LLMs can improve the readability to some extent. 
For example, ChatGPT and LLaMA2 rewrite `automotive industry' as `cars'. 
Claude2 translates `Ziegler-Natta catalysts' to `a specific type of catalyst'.
However, some academic domain-specific words (technical terminology) still exist. For example, the generated outputs still contain the phrases `high-density polyethylene' and `linear low-density polyethylene' from Claude2 and ChatGPT, `Ziegler-Natta catalysts' from ChatGPT and LLaMA2. 
Despite the catalyst terminology, LLaMA2 summarizes the academic text best, which is consistent with the experimental results of the simplicity metric (SARI) in Table~\ref{tab:main}.
In contrast, the fine-tuned generative language models (i.e., BART and Ours) do not introduce these technical terminologies. Fine-tuned BART generates some unrelated phrases (i.e. kinetic parameters), whereas our proposed model DSPT5 can still express the key idea from the source text that polymers are important and widely used in life and their manufacturing process is critical to industry. At the same time, there is no technical terminology in the output text generated by DSPT5. 
We did not include the output of fine-tuned T5 and FLAN-T5 as they yield similar outputs to the source text. We speculate that this is because there is no appropriate domain-specific prompt provided for fine-tuning T5 and FLAN-T5 with the limited training data.


\section{Human Evaluation}~\label{sec:appendix_human}

To provide a comprehensive assessment of the generated general-audience abstract, we conducted a human evaluation involving our proposed model DSPT5 and all other baseline models using three independent experts~\footnote{All judges have experience in scientific research and hold bachelor, master, and doctorate degree respectively.}.
Specifically, following a similar setting as~\cite{liu-etal-2024-sumsurvey, li-etal-2024-side, song-etal-2024-finesure, kew-etal-2023-bless, devaraj-etal-2021-paragraph}, our evaluation uses a random sample of 20 abstracts (10 in Computer Science and 10 in all other fields) from the test split VTechAGP considering the workload. Judges are presented with both the academic abstract and generated general-audience abstracts from seven models (DSPT5 and six baselines) for each data sample in a total of 140 abstracts. Using a 1-5 Likert scale, the judges are asked to rate the model output based on five criteria: comprehensiveness, layness, meaning preservation, conciseness, and fluency. 
Details are shown in guidelines, which clarify the meanings for each criterion in Figure~\ref{fig:guide} in the Appendix.
Table~\ref{tab:human} presents the average ratings from our human evaluation.

\section{Methodology Explanation Details}~\label{sec:explain}
In Sec.~\ref{sec:approach}, we discuss the contrastive loss infoNCE derived from different learning representations. Here we provide a detailed explanation.
To the textual level, $r$ represents a set of keywords (here a set of keywords are textual keywords, and then $r$ is the representation (embedding) of the several textual keywords). As we said, $r^{pos}$ and $r^{neg}$ are not sets of vectors. Instead, $r^{pos}$ is a vector and $r^{neg}$ is a vector. They are the vector representation of a set of textual keywords. 
If there are $m$ keywords in the golden label and keyBERT returns $n$ keywords (where $m<n$), we sort the keywords by confidence and select the top $m$ keywords. Then we concatenate the words for dot production.
For example, if there are two terms in the label, we select two positive keywords from KeyBERT. The selection is based on the ranking of the confidence value provided by KeyBERT.
We use T5 encoder, the inputs\_embeds are in [batch\_size, sequence\_length, hidden\_size], we concatenate embeddings in the sequence\_length dimension.
Although dot production is sensitive to the order of the keywords, currently we do a simple concatenation and we leave the works for exploring new ranking approaches to make the keyword extraction module extract closer or more relevant words to the golden label in the future as the main contributions for this paper is about the new dataset VTechAGP and implementations for benchmark models.

\section{Ethical Statements}
The VTechAGP dataset, derived from theses and dissertations available in VTechWorks,\footnote{Virginia Tech Electronic Theses and Dissertations: \url{https://hdl.handle.net/10919/5534}} is shared with a clear understanding of the ethical implications. 
Virginia Tech, as the institution where these ETDs were submitted, holds the right to distribute the content within the bounds of academic sharing and research purposes. 
The university's policy ensures that the distribution of such academic materials aligns with educational objectives and respects the authors' intellectual contributions.

Descriptive metadata, including titles, degree types, and departmental affiliations, is generally not subject to copyright restrictions. 
This metadata is typically viewed as factual information, which falls outside the scope of copyright protection. 
This understanding allows for the broad dissemination and analysis of metadata in academic and research contexts without infringing upon copyright laws.
The abstracts, however, may present a different scenario. 
While the university facilitates their distribution through VTechWorks, the copyright of the abstract text itself could potentially remain with the authors. 
Nevertheless, the use of abstracts in the VTechAGP can be justified under the principles of fair use, particularly given the dataset’s purpose in advancing scholarly research, education, and understanding. 
The dataset does not replicate the full ETDs but instead provides a means to study and analyze the structure and nature of academic abstracts themselves, thereby contributing to the field of information retrieval and digital libraries.

\begin{table*}[]
\centering
\caption{Examples of academic to general audience text paraphrasing comparison. The target text shows the gold label.}
\label{tab:case}
\begin{tabular}{ll}
\hline
Source Text & \makecell[l] {Polyolefins are one of the most widely used commodity polymers with applications in \\ films, packaging and automotive industry. The modeling of polymerization processes \\ producing polyolefins, including high-density polyethylene (HDPE), \\ polypropylene (PP), and linear low-density polyethylene (LLDPE) using Ziegler-Natta \\ catalysts with multiple active sites, is a complex and challenging task...} \\ \hline
Target Text & \makecell[l] {Almost everything we see around us from furniture, electronics to bottles, cars, etc. are \\ made fully or partially from plastic polymers. The two most popular polymers which \\ comprise almost two-thirds of polymer production globally are polyethylene (PE) and \\ polypropylene (PP), collectively known as polyolefins. Hence, the optimization of \\ polyolefin manufacturing processes with the aid of simulation models is critical and \\ profitable for chemical industry.} \\ \hline
BART        & \makecell[l]{Polyolefins are one of the most widely used commodity polymers with applications in \\ films, packaging and automotive industry. The modeling of polymerization processes \\ producing polyfins involves many processes and their kinetic parameters that have been \\ validated by scientific institutions and market models...} \\ \hline
Claude2      & \makecell[l]{Polyolefins are a category of plastics that are utilized across many industries for \\ products such as packaging materials, plastic bags, and automobile parts. Modeling the \\ manufacturing processes that create popular polyolefins like high-density polyethylene, \\ polypropylene, and linear low-density polyethylene, when using a specific type of \\ catalyst, involves intricate and difficult steps...} \\ \hline
ChatGPT     & \makecell[l]{Polyolefins are commonly used plastics found in items like films, packaging materials, \\ and parts for cars. Creating these plastics, such as high-density polyethylene (HDPE), \\ polypropylene (PP), and linear low-density polyethylene (LLDPE), through \\ polymerization processes involving Ziegler-Natta catalysts with multiple active sites \\ is a complex and challenging undertaking.} \\ \hline
LLaMA2      & \makecell[l]{Polyolefins are common plastics used in packaging, films, and car parts. Modeling how \\ these materials are made is complex and challenging, especially when using \\ Ziegler-Natta catalysts with multiple active sites...} \\ \hline
Ours        & \makecell[l]{The global population continues to grow, and with it, the demand for polyolefin will \\ further increase. Thus, there will never be a non-polymerization process, and thus, \\ many parts of the global production chain must be produced synthetically. Polyolefin \\ is the most widely used commodity in the automotive and aerospace industries. The \\ manufacturing of polyolefins is challenging since...} \\ \hline
\end{tabular}
\end{table*}

\begin{figure*}[htp] 
 \center{\includegraphics[height=18cm,width=\textwidth]{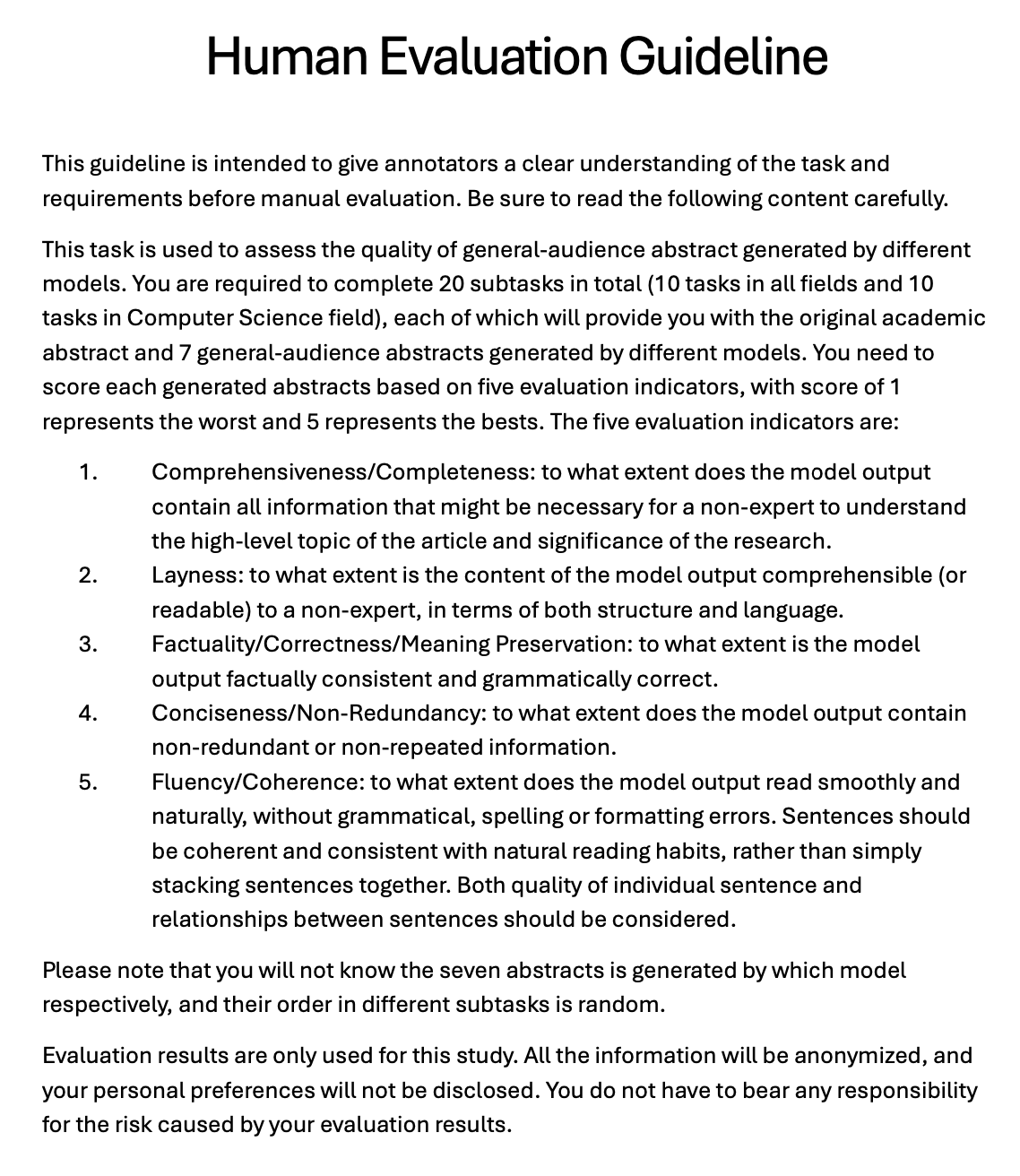}}
 \caption{\label{fig:guide} Human evaluation guideline.}
 \end{figure*}
\end{document}